%% file: article.tex
\documentclass[10pt,twocolumn,letterpaper]{article}

\usepackage{cvpr}

\usepackage{graphicx}
\usepackage{amsmath}
\usepackage{amssymb}
\usepackage{tikz}
\usepackage{subcaption}

\usetikzlibrary{arrows}

\usepackage[pagebackref=true,breaklinks=true,colorlinks,bookmarks=false,urlcolor=black]{hyperref}

\usepackage{float}

\cvprfinalcopy

\def\cvprPaperID{6964}

\begin{document}

\title{PifPaf: Composite Fields for Human Pose Estimation}

\author{Sven Kreiss, Lorenzo Bertoni, Alexandre Alahi\\
EPFL VITA lab\\
CH-1015 Lausanne\\
{\tt\small sven.kreiss@epfl.ch}
}

\maketitle


\begin{abstract}
    We propose a new bottom-up method for multi-person 2D human pose
    estimation that is particularly well suited for urban mobility such as self-driving cars
    and delivery robots. The new method, PifPaf, uses a Part Intensity Field (PIF) to
    localize body parts and a Part Association Field (PAF) to associate body parts with each other to form
    full human poses.
    Our method outperforms previous methods at low resolution and in crowded,
    cluttered and occluded scenes
    thanks to (i) our new composite field PAF encoding fine-grained information and (ii) the choice of Laplace loss for regressions which incorporates a notion of uncertainty.
    Our architecture is based on a fully
    convolutional, single-shot, box-free design.
    We perform on par with the existing
    state-of-the-art bottom-up method on the standard COCO keypoint task
    and produce state-of-the-art results on a modified COCO keypoint task for
    the transportation domain.
\end{abstract}

\section{Introduction}

Tremendous progress has been made in estimating human poses ``in the wild"  driven by popular data collection campaigns~\cite{andriluka14cvpr,lin2014microsoft}. Yet, when it comes to the ``transportation domain" such as for self-driving cars or social robots, we are still far from matching an acceptable level of accuracy. While a pose estimate is not the final goal, it is an effective low dimensional and interpretable representation of humans to detect critical actions early enough for autonomous navigation systems  (\eg, detecting pedestrians who intend to cross the street). Consequently, the further away a human pose can be detected, the safer an autonomous system will be. This directly relates to pushing the limits on the minimum resolution needed to perceive human poses.

\begin{figure}
  \centering
  \includegraphics[width=\linewidth]{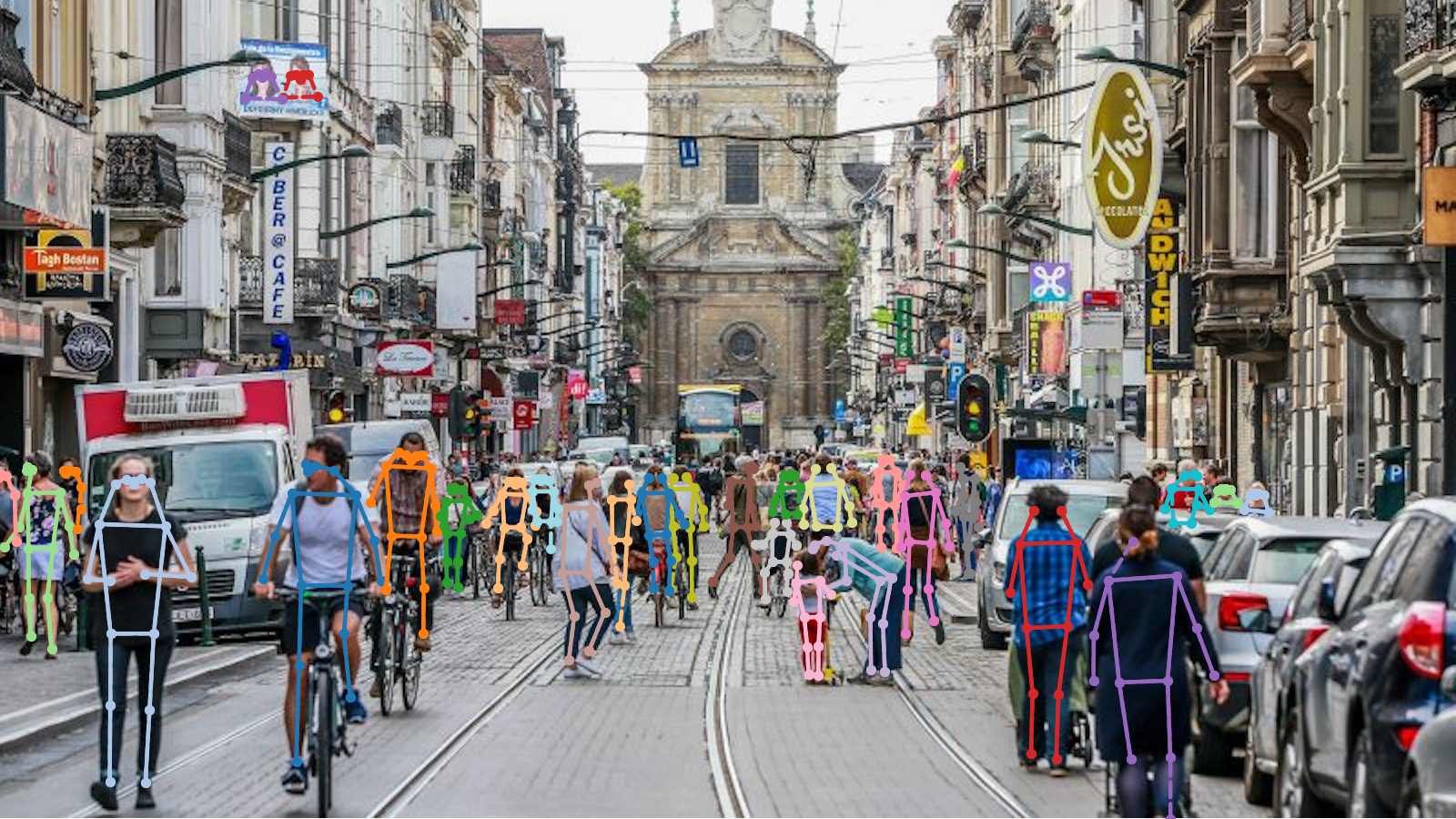}

  \caption{We want to estimate human 2D poses in the transportation domain where autonomous navigation systems operate in crowded scenes. Humans occupy small portion of the images and could partially occlude each other. We show the output of our PifPaf method with colored segments.}
  \label{fig:pull}
\end{figure}

In this work, we tackle the well established multi-person 2D human pose estimation problem given a single input image. We specifically address challenges that arise in autonomous navigation settings as illustrated in Figure~\ref{fig:pull}:  (i) wide viewing angle with limited resolution on humans, \ie, a height of 30-90 pixels, and (ii) high density crowds where pedestrians occlude each other. Naturally, we aim for high recall and precision.

Although pose estimation has been studied before the deep learning era, a significant cornerstone is the work of OpenPose~\cite{partsaffinityfields}, followed by Mask R-CNN~\cite{he2017mask}. The former is a bottom-up approach (detecting joints without a person detector), and the latter is a top-down one (using a person detector first and outputting joints within the detected bounding boxes). While the performance of these methods is stunning on high enough resolution images, they perform poorly in the limited resolution regime, as well as in dense crowds where humans partially occlude each other.

In this paper, we propose to extend the notion of fields in pose
estimation~\cite{partsaffinityfields} to go beyond scalar and vector fields to \emph{composite}
fields. We introduce a new neural network architecture with two head networks. For each
body part or joint, one head network predicts the confidence score, the precise
location and the size of this joint, which we call a Part Intensity Field (PIF) and
which is similar to the fused part confidence map in~\cite{personlab}. The other head
network predicts associations between parts, called the Part Association Field (PAF),
which is of a new composite structure. Our encoding scheme has the capacity to store
fine-grained information on low resolution activation maps.
The precise regression to joint locations is critical, and we use a Laplace-based $L_1$ loss~\cite{kendall2017uncertainties} instead of the vanilla $L_1$ loss~\cite{he2017mask}.
Our experiments show that we outperform both bottom-up and established top-down methods on low resolution images while performing on par on higher resolutions. The software is open source and available
online\footnote{\url{https://github.com/vita-epfl/openpifpaf}}.

\section{Related Work}

Over the past years, state-of-the-art methods for pose estimation are based on
Convolutional Neural Networks~\cite{he2017mask,partsaffinityfields,newell2017associative,personlab}.
They outperform traditional methods based on pictorial
structures~\cite{felzenszwalb2005pictorial,dantone2013human,eichner20122d} and
deformable part models~\cite{felzenszwalb2010object}. The deep learning tsunami
started with DeepPose~\cite{toshev2014deeppose} that uses a cascade of convolutional
networks for full-body pose estimation. Then, instead of predicting absolute human
joint locations, some works refine pose estimates by predicting error
feedback (\ie, corrections) at each iteration~\cite{carreira2015human,haque2016towards} or using a
human pose refinement network to exploit dependencies between input and output
spaces~\cite{Fieraru2018LearningTR}. There is now an arms race towards proposing
alternative neural network architectures: from convolutional pose
machines~\cite{wei2016convolutional}, stacked hourglass
networks~\cite{newell2016stacked, Luo2019MultiPersonPE}, to recurrent
networks~\cite{belagiannis2017recurrent}, and voting schemes
such as~\cite{lifshitz2016human}.

All these approaches for human pose estimation can be grouped into bottom-up and
top-down methods. The former one estimates each body joint first and then groups
them to form a unique pose. The latter one runs a person detector first and estimates
body joints within the detected bounding boxes.

\paragraph{Top-down methods.} Examples of top-down methods are
PoseNet~\cite{papandreou2017towards}, RMPE~\cite{Fang2017RMPERM},
CFN~\cite{Huang2017ACN}, Mask R-CNN~\cite{he2017mask,Detectron2018} and more recently
CPN~\cite{chen2018cascaded} and MSRA~\cite{xiao2018simple}. These methods profit from
advances in person detectors and vast amounts of labeled bounding boxes
for people. The ability to leverage that data turns the requirement of a
person detector into an advantage. Notably, Mask R-CNN treats keypoint detections as an
instance segmentation task. During training, for every independent keypoint, the target is
transformed to a binary mask containing a single foreground pixel. In general, top-down
methods are effective but struggle when person bounding boxes overlap.

\paragraph{Bottom-up methods.} Bottom-up methods include the pioneering work by Pishchulin with
DeepCut~\cite{pishchulin2016deepcut}
and Insafutdinov with DeeperCut~\cite{insafutdinov2016deepercut}. They solve the part
association with an integer linear program which results in processing
times for a single image of the order of hours.
Later works accelerate the prediction time~\cite{chen2018deeplab} and broaden
the applications to track animal behavior~\cite{mathis2018deeplabcut}.
Other methods drastically reduce prediction time by using greedy decoders in combination with
additional tools as in
Part Affinity Fields~\cite{partsaffinityfields},
Associative Embedding~\cite{newell2017associative} and
PersonLab~\cite{personlab}. Recently, MultiPoseNet~\cite{Kocabas2018MultiPoseNetFM}
develops a multi-task learning architecture combining detection, segmentation and pose
estimation for people.

Other intermediate representations have been build on top of
2D pose estimates in the image plane including
3D pose estimates~\cite{martinez2017simple},
human pose estimation in videos~\cite{pfister2015flowing} and
dense pose estimation~\cite{guler2018densepose} that would all profit
from improved 2D pose estimates.

\section{Method}

\begin{figure*}[t]
  \centering
  \includegraphics[width=0.75\linewidth]{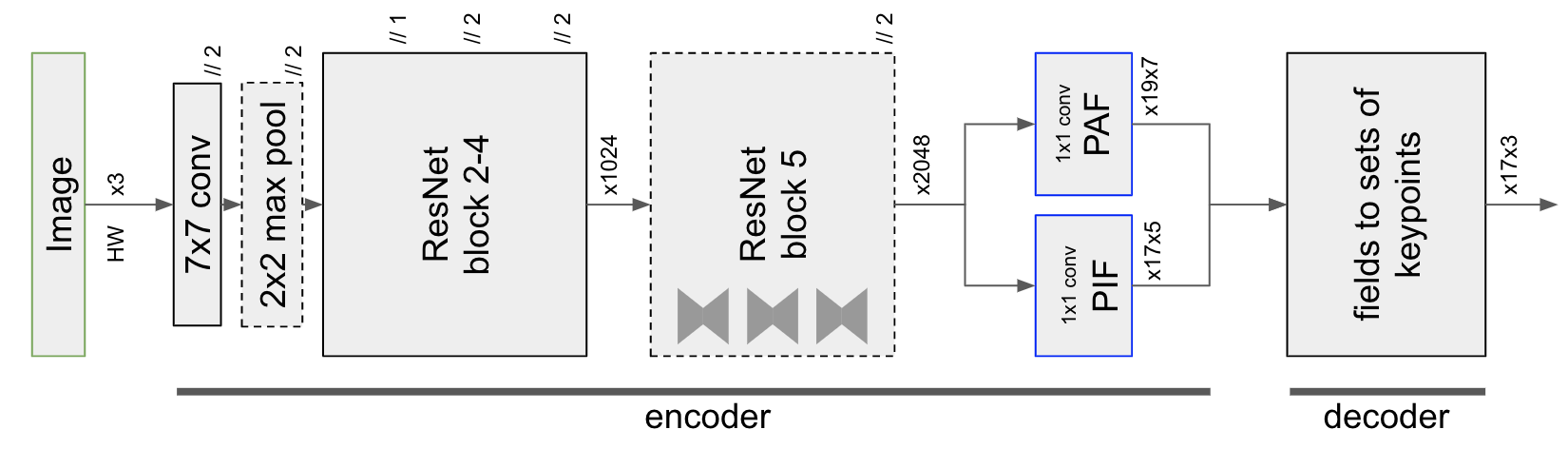}
  \includegraphics[width=0.24\linewidth,trim=0 -11em 0 0]{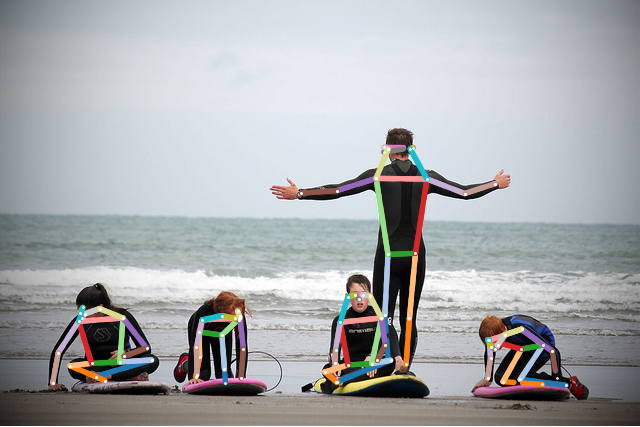}
  \caption[Model architecture]{
    Model architecture. The input is an image of size $(H,W)$ with three color
    channels, indicated by ``x3''.
    The neural network based encoder produces PIF and PAF fields with
    $17\times5$ and $19\times7$ channels. An operation with stride two is indicated by ``//2''.
    The decoder is a program that converts PIF and PAF fields into pose
    estimates containing 17 joints each. Each joint is represented by an $x$ and $y$
    coordinate and a confidence score.
  }
  \label{fig:model}
\end{figure*}

The goal of our method is to estimate human poses in crowded images. We address
challenges related to low-resolution
and partially occluded pedestrians. Top-down methods particularly struggle when
pedestrians are occluded by other pedestrians where bounding boxes clash.
Previous bottom-up methods are bounding box free but still contain a coarse feature
map for localization. Our method is free of any grid-based constraint on the
spatial localization of the joints and
has the capacity to estimate multiple poses occluding each other.

Figure~\ref{fig:model} presents our overall model. It is a shared
ResNet~\cite{he2016deep} base network with two head networks:
one head network predicts a confidence, precise location and size of a joint, which we
call a Part Intensity Field (PIF), and the other head network predicts associations between parts, called the Part Association Field (PAF). We refer to our method as \textit{PifPaf}.

Before describing each head network in detail, we briefly define our field notation.

\subsection{Field Notation}

Fields are a useful tool to reason about structure on top of images.
The notion of composite fields
directly motivates our proposed Part Association Fields.

We will use $i,j$ to enumerate spatially the output locations of the neural
network and $x,y$ for real-valued coordinates. A field is denoted with
$\textbf{f}^{ij}$ over the domain $(i, j) \in \mathbb{Z}_+^2$ and can have
as codomain (the values of the field) scalars, vectors or composites.
For example, the composite of a scalar field  $\textbf{s}^{ij}$ and a vector field $\textbf{v}^{ij}$
can be represented as $\{s, v_x, v_y\}$
which is equivalent to ``overlaying'' a
confidence map with a vector field.

\subsection{Part Intensity Fields}

The Part Intensity Fields (PIF) detect and precisely localize body parts.
The fusion of a confidence map with a regression
for keypoint detection was introduced in~\cite{papandreou2017towards}.
Here, we recap this technique in the language of composite fields and add
a scale $\sigma$ as a new component to form our PIF field.

PIF have composite structure. They are
composed of a scalar component for confidence, a vector component that points
to the closest body part of the particular type and another scalar component for the size of the joint.  More formally, at every output location $(i, j)$, a PIF predicts a confidence~$c$,
a vector~$(x, y)$ with spread~$b$ (details in Section~\ref{sec:adaptive-r}) and a scale~$\sigma$ and can be written as
$\textbf{p}^{ij} = \{p^{ij}_c, p^{ij}_x, p^{ij}_y, p^{ij}_b, p^{ij}_\sigma\}$.

\begin{figure*}
  \centering
  \begin{subfigure}[b]{0.33\linewidth}
    \centering
    \includegraphics[width=\linewidth,trim=0 0 0 0]{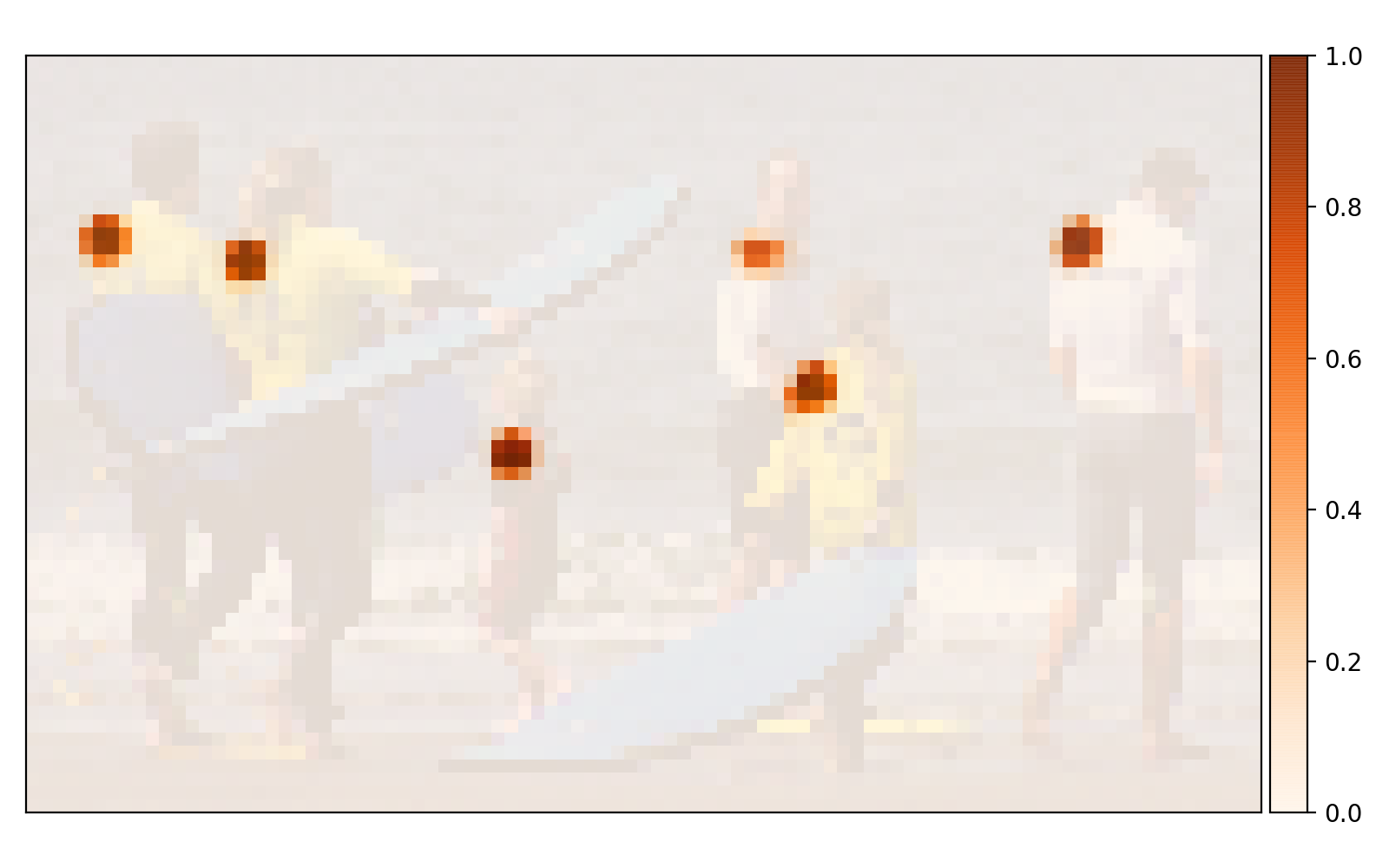}
    \caption{}
    \label{fig:pif-confidence}
  \end{subfigure}
  \begin{subfigure}[b]{0.315\linewidth}
    \centering
    \includegraphics[width=\linewidth,trim=0 -2em 0 0]{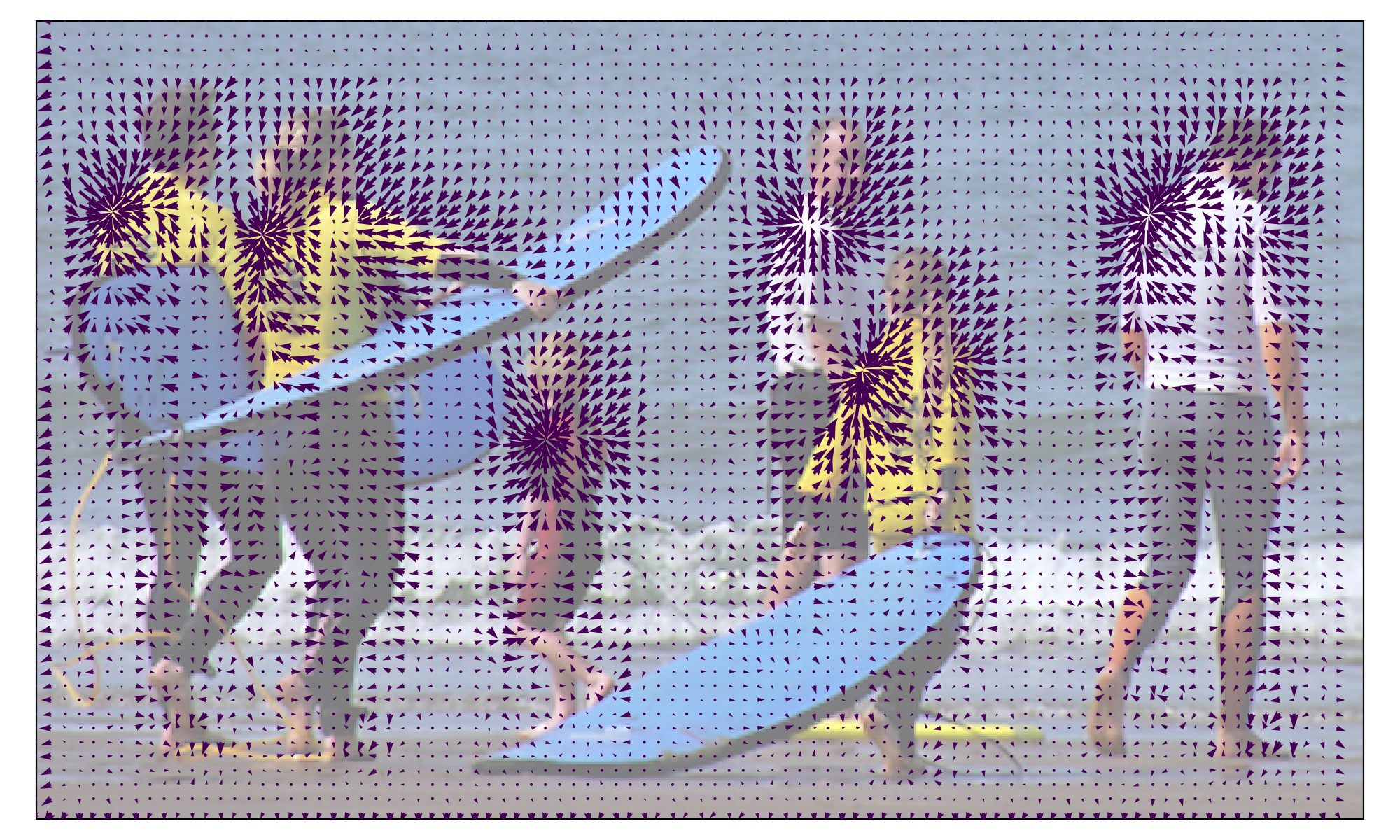}
    \caption{}
    \label{fig:pif-vector}
  \end{subfigure}
  \begin{subfigure}[b]{0.33\linewidth}
    \centering
    \includegraphics[width=\linewidth,trim=0 0 0 0]{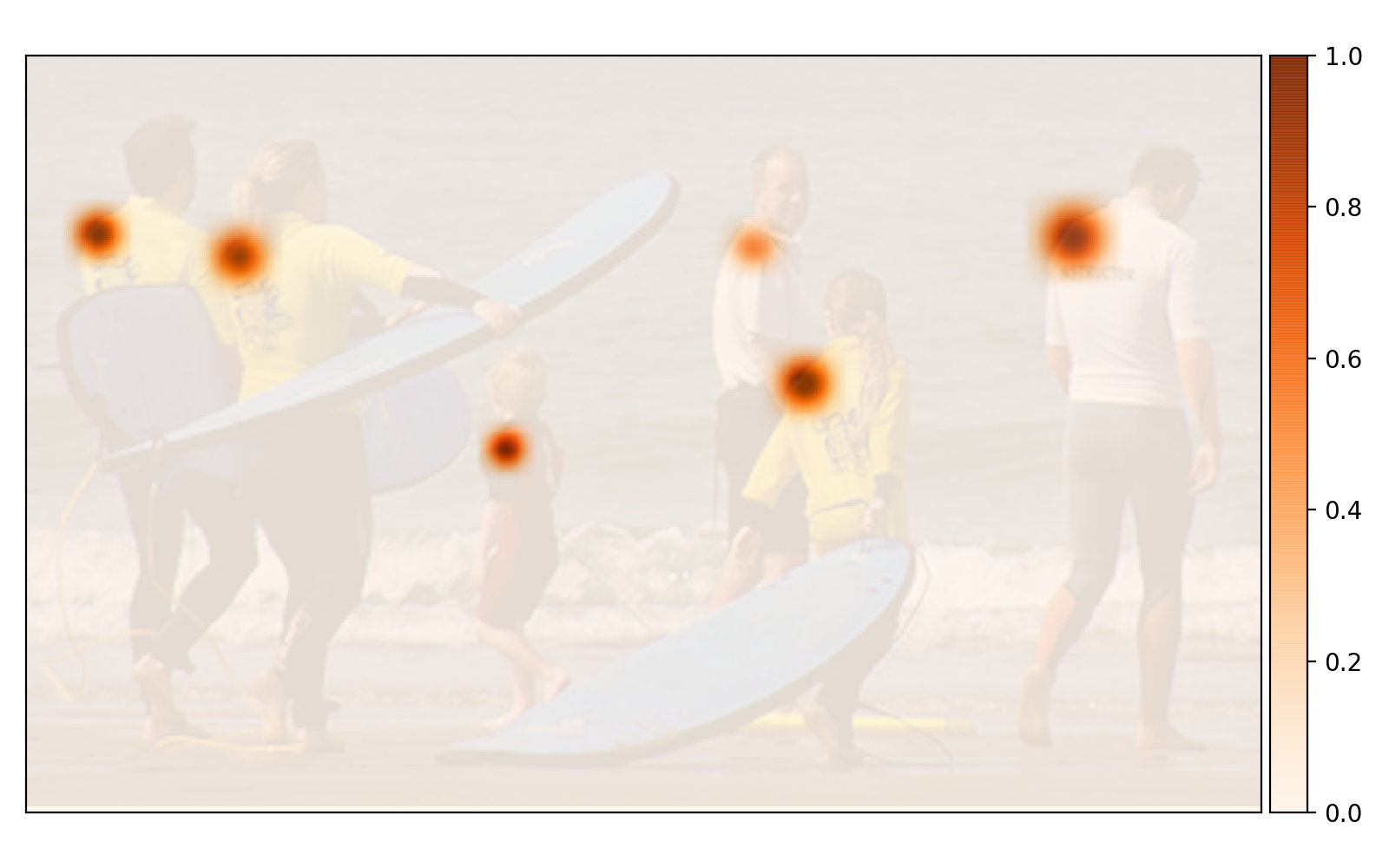}
    \caption{}
    \label{fig:pif-fused}
  \end{subfigure}
  \caption{
    Visualizing the components of the PIF for the left shoulder. This is one of the 17 composite PIF. The confidence map is shown in~(\subref{fig:pif-confidence})
    and the vector field is shown in~(\subref{fig:pif-vector}). The fused confidence, vector and scale components are shown
    in~(\subref{fig:pif-fused}).
  }
  \label{fig:pif2}
\end{figure*}

The confidence map of a PIF is very coarse. Figure~\ref{fig:pif-confidence} shows
a confidence map for the left shoulders
for an example image. To improve the
localization of this confidence map, we fuse it with the vectorial part of the
PIF shown in
Figure~\ref{fig:pif-vector} into a high resolution confidence map.
We create this high resolution part confidence map $f(x, y)$ with a convolution
of an unnormalized Gaussian kernel $\mathcal{N}$ with width $p_\sigma$ over the regressed targets from
the Part Intensity Field weighted by its confidence~$p_c$:
\begin{equation}
  \label{eq:pifhr}
  f(x, y) = \sum_{ij} p^{ij}_c \; \mathcal{N} (x, y | p^{ij}_x, p^{ij}_y, p^{ij}_\sigma) \;\;\; .
\end{equation}
This equation emphasizes the grid-free nature of the localization. The spatial
extent $\sigma$ of a joint is learned as part of the field.
An example is shown in Figure~\ref{fig:pif-fused}.
The resulting map of highly localized
joints is used to seed the pose generation and to score the location of newly
proposed joints.

\subsection{Part Association Fields}

Associating joints into multiple poses is challenging in crowded scenes where people
partially occlude each other. Especially two step processes -- top-down methods --
struggle in this situation: first they detect person bounding boxes and then they attempt
to find one joint-type for each bounding box. Bottom-up methods are bounding box
free and therefore do not suffer from the clashing bounding box problem.

\begin{figure}
  \centering
  \begin{subfigure}{0.49\linewidth}
    \centering
    \input{images/offset_grid.tex}
    \caption{mid-range offsets}
    \label{fig:mid-range-offsets}
  \end{subfigure}
  \begin{subfigure}{0.49\linewidth}
    \centering
    \input{images/paf_grid.tex}
    \caption{Part Association Field}
    \label{fig:parts-association-fields}
  \end{subfigure}
  \caption{Illustrating the difference between
           PersonLab's mid-range offsets (a) and
           Part Association Fields (b) on a feature map grid.
           Blue circles represent joints and confidences are marked in green.
           Mid-range offsets (a) have their origins at the center of feature map cells.
           Part Association Fields (b) have floating point precision of their origins.}
  \label{fig:mid-range-and-parts-association-fields}
\end{figure}
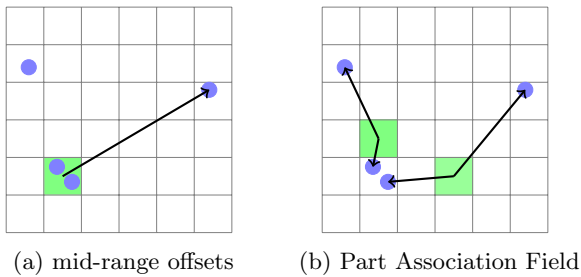

\begin{figure*}
  \centering
  \begin{subfigure}[b]{0.49\linewidth}
    \centering

    \includegraphics[width=\linewidth]{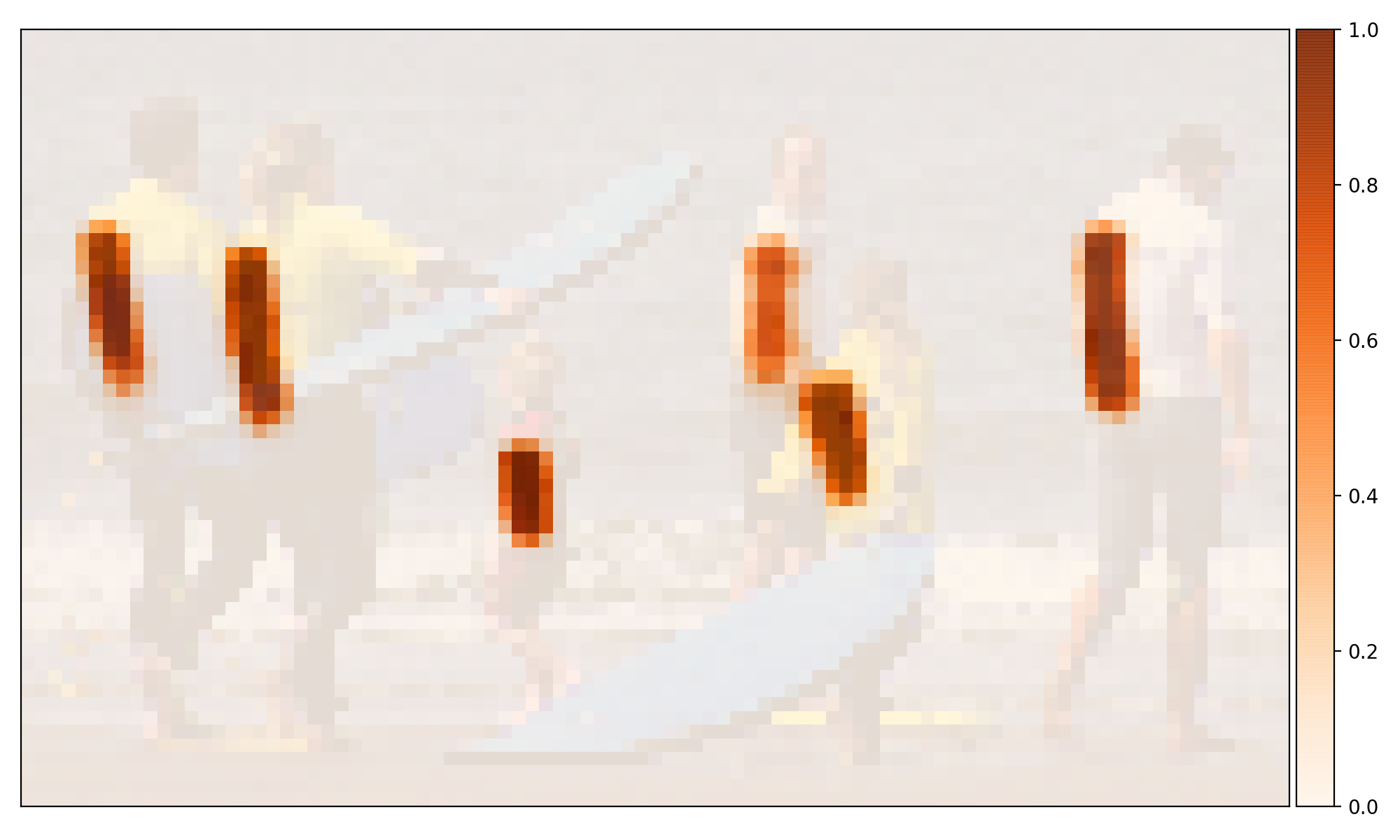}
    \caption{}
    \label{fig:paf2-intensity}
  \end{subfigure}
  \begin{subfigure}[b]{0.49\linewidth}
    \centering

    \includegraphics[width=\linewidth]{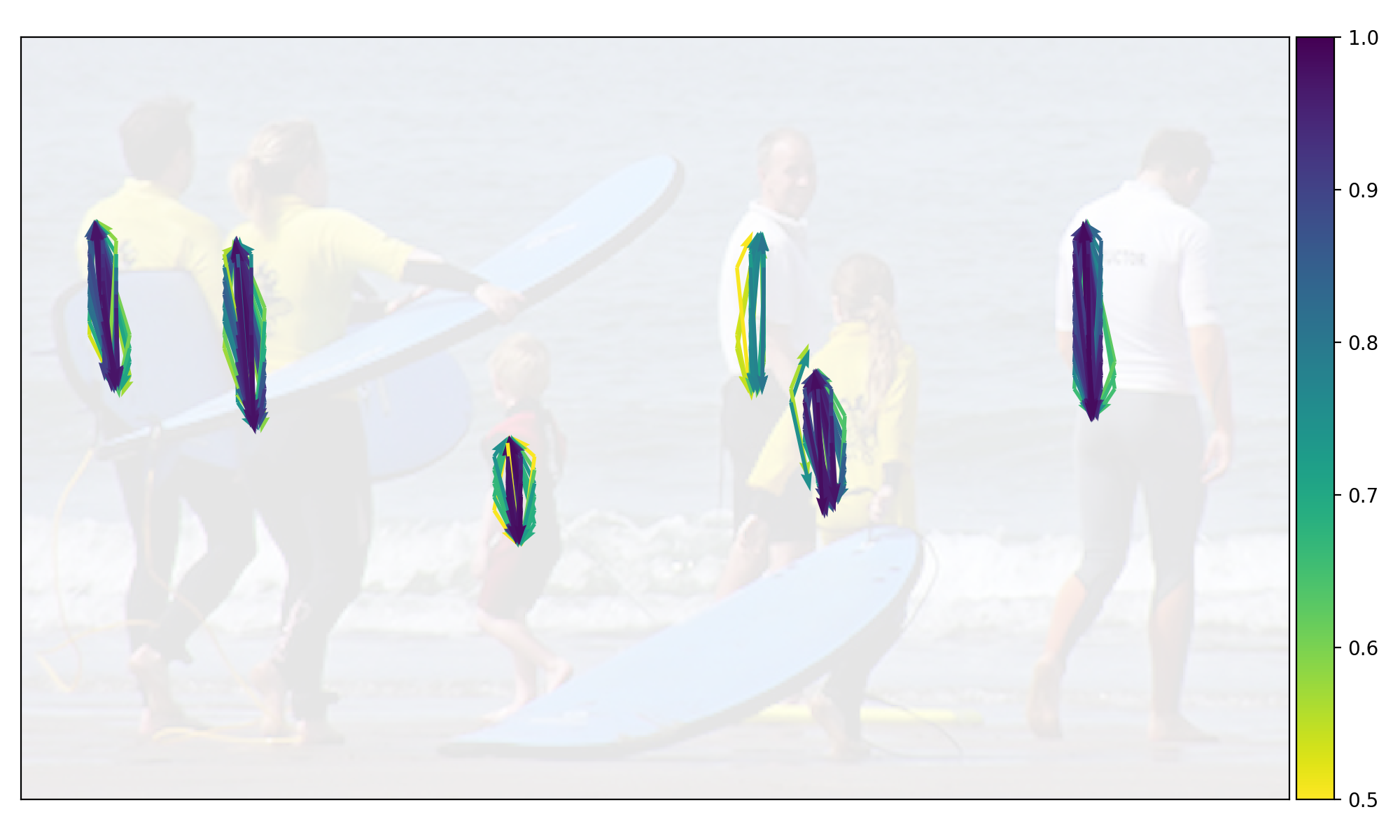}
    \caption{}
    \label{fig:paf2-breakdown}
  \end{subfigure}
  \caption{Visualizing the components of the PAF that associates left shoulder with left hip. This is one of the 19 PAF. Every location of the feature map is the origin of two vectors which point to the shoulders and hips to associate. The confidence of associations $\textbf{a}_c$ is shown at their origin in (\subref{fig:paf2-intensity}) and the vector components for $\textbf{a}_c > 0.5$ are shown in (\subref{fig:paf2-breakdown}).}
  \label{fig:paf2}
\end{figure*}

We propose bottom-up Part Association Fields~(PAF) to connect joint locations together into poses.
An illustration of the PAF scheme is shown in
Figure~\ref{fig:mid-range-and-parts-association-fields}.
At every output location,
PAFs predict a confidence, two vectors to the two parts this association is
connecting and two widths $b$ (details in Section~\ref{sec:adaptive-r}) for the spatial precisions of the regressions.
PAFs are represented with
$\textbf{a}^{ij} = \{a^{ij}_c, a^{ij}_{x1}, a^{ij}_{y1}, a^{ij}_{b1}, a^{ij}_{x2}, a^{ij}_{y2}, a^{ij}_{b2}\}$.
Visualizations of the associations between left shoulders
and left hips are shown in Figure~\ref{fig:paf2}.

Both endpoints are localized with regressions that do not suffer from discretizations
as they occur in grid-based methods. This helps to resolve joint locations of close-by
persons precisely and to resolve them into distinct annotations.

There are 19 connections for the person class in the COCO dataset each
connecting two types of joints; \textit{e.g.}, there is a right-knee-to-right-ankle
association. The algorithm to construct the PAF components at a particular
feature map location consists of two steps. First, find the closest joint of either of the two
types which determines one of the vector components. Second, the ground truth pose
determines the other vector component to represent the association. The second joint
is not necessarily the closest one and can be far away.

During training, the components of the field have to point to the parts
that should be associated. Similar to how an $x$ component of a vector field
always has to point to the same target as the $y$ component, the components of
the PAF field have to point to the same association of parts.

\subsection{Adaptive Regression Loss}
\label{sec:adaptive-r}

Human pose estimation algorithms tend to struggle with the diversity of scales that
a human pose can have in an image. While a localization error for the joint of
a large person can be minor, that same absolute error might be a major mistake
for a small person. We use an $L_1$-type loss to train regressive outputs.
We improve the localization ability of the network by injecting
a scale dependence into that regression loss with the
SmoothL1~\cite{girshick2015fast} or Laplace loss~\cite{kendall2017uncertainties}.

The SmoothL1 loss allows to tune the radius $r^\textrm{smooth}$ around the
origin where it produces softer gradients.
For a person
instance bounding box area of $A_i$ and keypoint size of $\sigma_k$,
$r^\textrm{smooth}_{i,k}$ can be set proportionally to $\sqrt{A_i} \sigma_k$
which we study in Table~\ref{tab:ablation}.

The Laplace loss is another $L_1$-type loss that is attenuated via the predicted
spread $b$:
\begin{equation}
  L = |x-\mu|/b + \log(2b) \;\;\; .
\end{equation}
It is independent of any estimates of
$A_i$ and $\sigma_k$ and we use it for all vectorial components.

\subsection{Greedy Decoding}

Decoding is the process of converting the output feature maps of a neural
network into sets of 17~coordinates that make human pose estimates.
Our process is similar to the
fast greedy decoding used in~\cite{personlab}.

A new pose is seeded by PIF vectors with the highest values in the
high resolution confidence map $f(x,y)$ defined in equation~\ref{eq:pifhr}.
Starting from a seed, connections to other joints are added with the help
of PAF fields. The algorithm is fast and greedy. Once a connection to a new
joint has been made, this decision is final.

Multiple PAF associations can form connections between the current and the
next joint.
Given the location of a starting
joint $\vec{x}$, the scores $s$ of PAF associations $\textbf{a}$ are calculated with
\begin{equation}
  s(\textbf{a}, \vec{x}) = a_c \;\; \exp\left(-\frac{||\vec{x} - \vec{a}_1||_2}{b_1}\right) \;\; f_2(a_{x2}, a_{y2})
\end{equation}
which takes into account the confidence in this connection $a_c$, the distance to the first
vector's location calibrated with the two-tailed Laplace distribution probability and the high resolution
part confidence at the second vector's target location $f_2$.
To confirm the proposed position of the new joint, we run reverse matching.
This process is repeated until a full pose is obtained.
We apply non-maximum suppression at the keypoint level as in~\cite{personlab}.
The suppression radius is dynamic and based on the predicted scale component of
the PIF field.
We do not refine any fields neither during training nor test time.

\section{Experiments}

\begin{table*}
  \centering
  \begin{tabular}{|l|c c c c c|c c c c c|}
    \hline
                           & AP & AP$^{0.50}$ & AP$^{0.75}$ & AP$^M$ & AP$^L$ & AR & AR$^{0.50}$ & AR$^{0.75}$ & AR$^M$ & AR$^L$ \\
    \hline\hline
    Mask R-CNN$^{*}$~\cite{he2017mask}        & 41.6 & 68.1 & 42.5 & 28.2 & 59.8 & 49.0 & \textbf{76.0} & 50.0 & 35.6 & 67.5 \\
    \hline
    OpenPose~\cite{partsaffinityfields}       & 37.6 & 62.5 & 37.2 & 25.0 & 55.3 & 43.9 & 65.3 & 44.9 & 26.7 & 67.5 \\

    \textbf{PifPaf} (ours)
    & \textbf{50.0} & \textbf{73.5} & \textbf{52.9} & \textbf{35.9} & \textbf{69.7} & \textbf{55.0} & \textbf{76.0} & \textbf{57.9} & \textbf{39.4} & \textbf{76.4} \\
    \hline
  \end{tabular}
  \caption{Applying pose estimation to low resolution images with the long side equal to $321\,$px for top-down (top part) and bottom-up (bottom part) methods.
           For the Mask R-CNN and OpenPose
           reference values, we ran the implementations by~\cite{tseng2018detectron, 8486591}
           modified to enforce the maximum image side length.
           $^{*}$Mask R-CNN was retrained for low resolution.
           The PifPaf result is based on a ResNet50 backbone.}
  \label{tab:low-res}
\end{table*}

Cameras in self-driving cars have a wide field of view and have to resolve
small instances of pedestrians within that field of view.
We want to emulate that small pixel-height distribution of pedestrians with a
publicly available dataset and evaluation protocol for human pose estimation.

In addition, and to demonstrate the
broad applicability of our method, we also investigate pose estimation in the context of the person re-identification
task (Re-Id) -- that is,
given an image of a person, identify that person in other images.
Some prior work has used part-based or region-based models~\cite{zhao2017spindle,Cheng_2016_CVPR,wu2015viewpoint}
that would profit from quality pose estimates.

\paragraph{Datasets.}

We quantitatively evaluate our proposed method, PifPaf, on the COCO keypoint task~\cite{lin2014microsoft} for people in
low resolution images. Starting from the original COCO dataset, we constrain the maximum image side length to 321 pixels to emulate a crop of a 4k camera. We obtain person bounding boxes that are
$66 \pm 65\,$px high. The COCO metrics contain a
breakdown for medium-sized humans under AP$^M$ and AR$^M$ that have bounding
box area in the original image between between $(32\,\textrm{px})^2$ and
$(96\,\textrm{px})^2$. After resizing for low resolution, this corresponds
to bounding boxes of height $44 \pm 19\,$px.

We qualitatively study the performance of our meth\-od on images captured by
self-driving cars as well as random crowded scenarios. We use the recently
released \textit{nuScenes} dataset~\cite{nuscenes}. Since labels and
evaluation protocols are not yet available we qualitatively study the results.

\begin{figure*}
  \vspace{1em}
  \centering
  \includegraphics[width=0.097\linewidth]{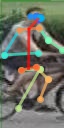}%
  \includegraphics[width=0.097\linewidth]{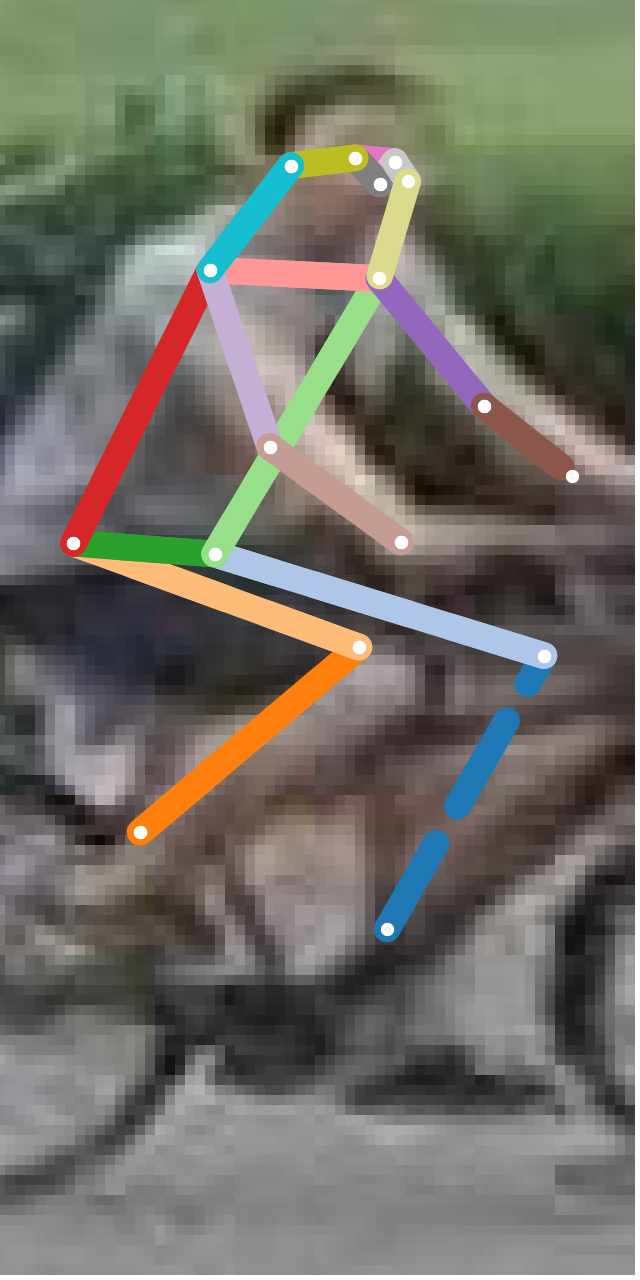}
  \includegraphics[width=0.097\linewidth]{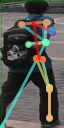}%
  \includegraphics[width=0.097\linewidth]{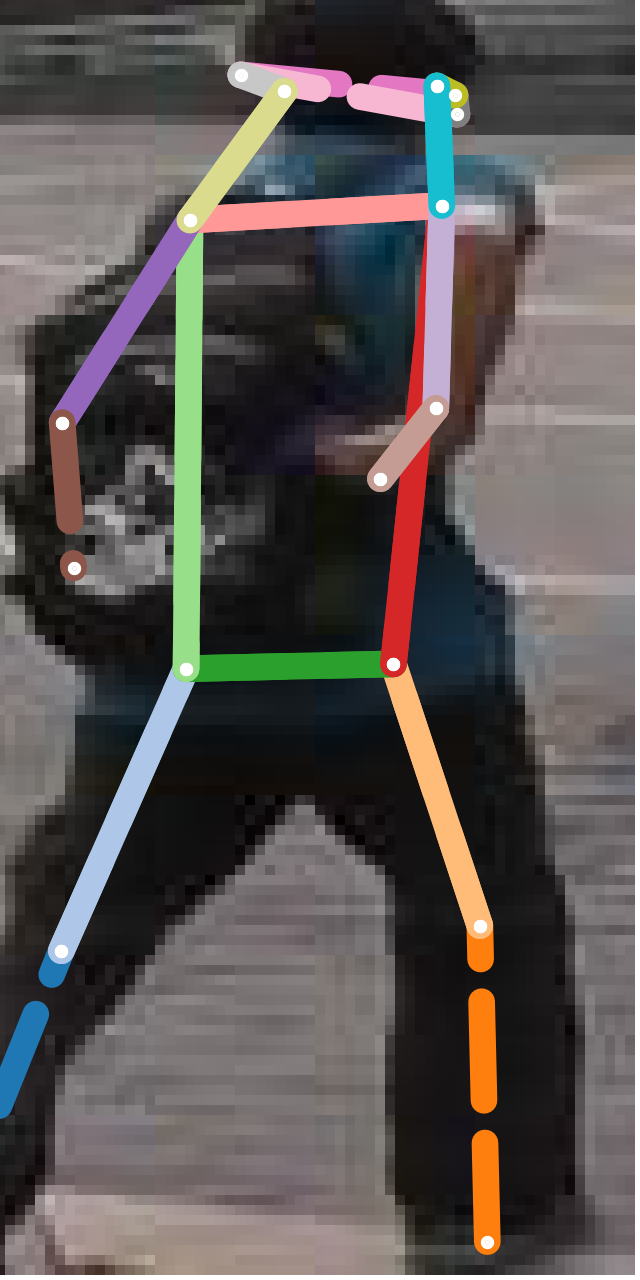}
  \includegraphics[width=0.097\linewidth]{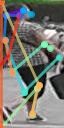}%
  \includegraphics[width=0.097\linewidth]{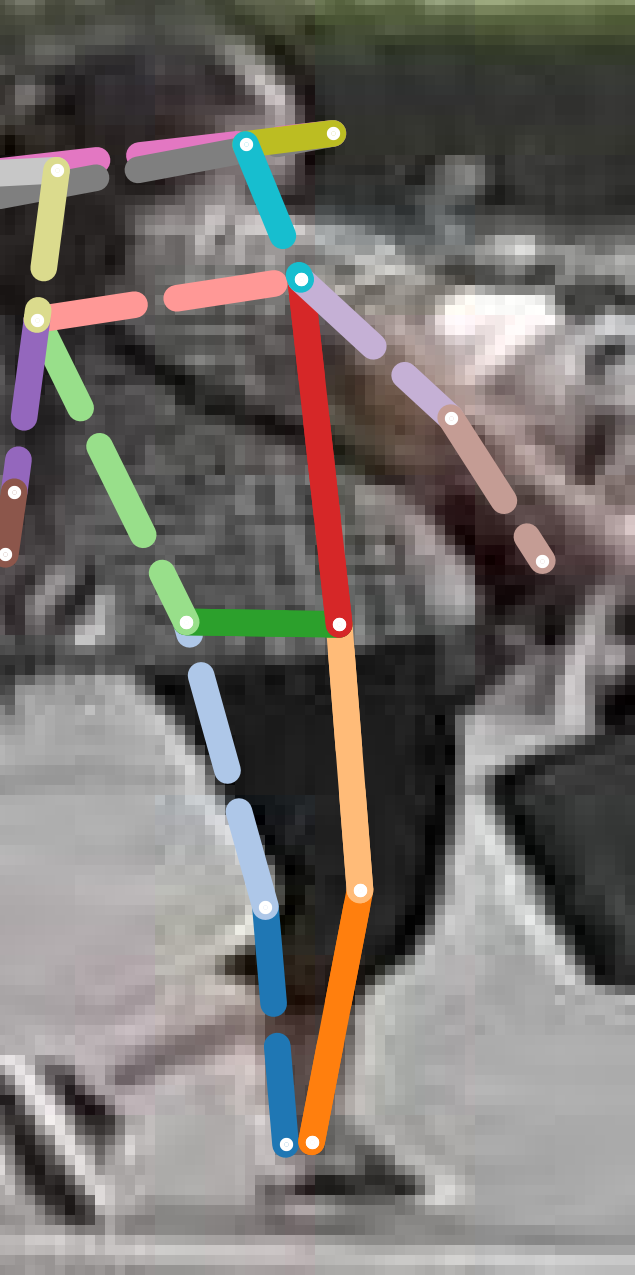}
  \includegraphics[width=0.097\linewidth]{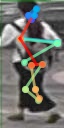}%
  \includegraphics[width=0.097\linewidth]{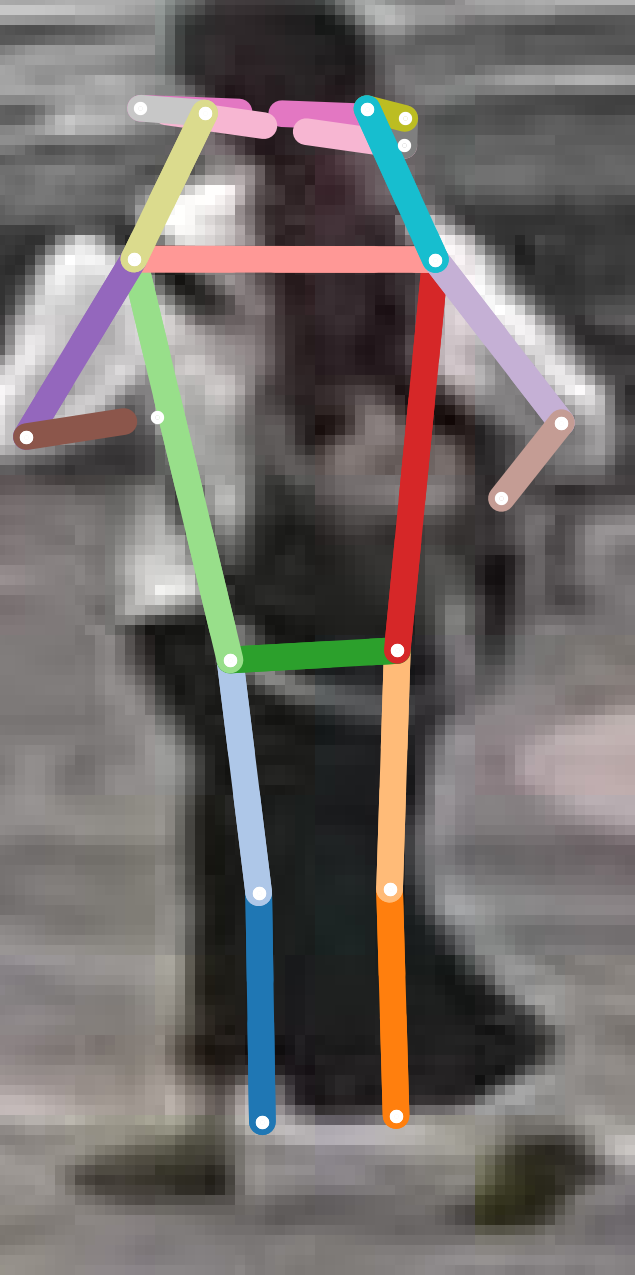}
  \includegraphics[width=0.097\linewidth]{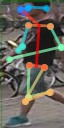}%
  \includegraphics[width=0.097\linewidth]{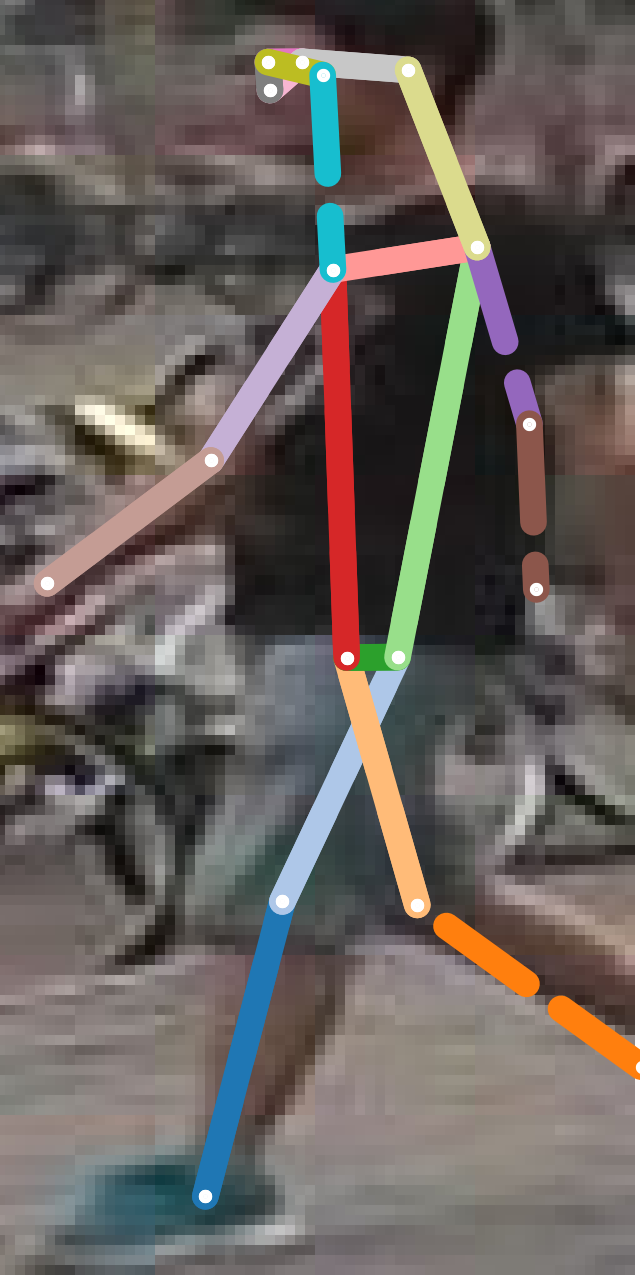}
  \caption{A selection of images from the Market-1501~\cite{zheng2015scalable} dataset.
           The left image is the output from
           Mask R-CNN. To improve the Mask R-CNN result, we forced it to
           predict exactly one pose in a bounding box that spans the entire image.
           The right image is the output of our PifPaf method that was not
           constrained to one person and could have chosen to output none
           or multiple poses, which is a harder task.}
  \label{fig:market1501}
\end{figure*}

In the context of Re-Id, we investigate the popular and publicly available Market-1501
dataset~\cite{zheng2015scalable}. It consists of $64 \times 128$ pixel crops of
pedestrians.
We apply the same model that we trained on COCO data.
Figure~\ref{fig:market1501} qualitatively compares extracted poses
from Mask R-CNN~\cite{he2017mask} with our proposed method. The comparison shows
a clear improvement of the poses extracted with our PifPaf method.

Performance on higher resolution images is not the focus of this paper, however
other methods are optimized for full resolution COCO images and therefore we also
show our results and comparisons for high resolution COCO poses.

\paragraph{Evaluation.}

The COCO keypoint detection task is evaluated like an object detection task, with the core
metrics being variants of average precision (AP) and average recall (AR) thresholded at an
object keypoint similarity (OKS)~\cite{lin2014microsoft}.
COCO assumes a fixed ratio of keypoint size to bounding box area per keypoint type to
define OKS.
For each image, pose estimators have to provide the 17 keypoint locations per pose and a score for each pose. Only the top 20 scoring poses are considered for evaluation.

\paragraph{Implementation details.}

All our models are based on Imagenet pretrained base networks followed
by custom, multiple head sub-networks. Specifically, we use the 64115 images in the 2017
COCO training set that have a person annotation for training. Our validation is done on
the 2017 COCO validation set of 5000 images. The base networks are modified
ResNet50/101/152 networks. The head networks are single-layer 1x1 sub-pixel
convolutions~\cite{shi2016real} that double the spatial resolution.
The confidence component of a field is normalized with a sigmoid non-linearity.

The base network has various modification options. The strides of the input convolution
and the input max-pooling operation can be changed.
It is also possible to remove the max-pooling operation in the input block and the entire
last block. The default modification used here is to remove the max-pool layer from the input
block.

We apply only few and weak data
augmentations. To create uniform batches, we crop images to squares where the side of the
square is between 95\% and 100\% of the short edge of the image and the location is
chosen randomly. These are large crops to keep as much of the training data as possible.
Half of the time the entire image is used un-cropped and bars are added
to make it square.
The subsequent resizing uses bicubic interpolation.
Training images and annotations are randomly horizontally flipped.

The components of the fields that form confidence maps are trained with independent binary
cross entropy losses. We use $L_1$ losses for the scale components of the PIF fields and use
Laplace losses for all vectorial components.

During training, we fix the running statistics of the Batch Normalization
operations~\cite{batchnormalization} to their pretrained values~\cite{personlab}.
We use the SGD optimizer with a learning rate of $10^{-3}$, momentum of 0.95, batch size of 8 and no weight decay.
We employ model averaging to extract stable models for validation. At each optimization
step, we update an exponentially weighted version of the model parameters. Our decay
constant is $10^{-3}$.
The training time for 75 epochs of ResNet101 on two GTX1080Ti is approximately 95~hours.

\paragraph{Baselines.} We compare our proposed PifPaf method against the reproducible
state-of-the-art bottom-up OpenPose~\cite{partsaffinityfields} and top-down Mask
R-CNN~\cite{he2017mask} methods. While our goal is to outperform bottom-up approaches, we
still report results of a top-down approach to evaluate the strength of our method. Since
this is an emulation of small humans within
a much larger image, we modified existing methods to prevent upscaling of small
images.

\paragraph{Results.}
Table~\ref{tab:low-res} presents our quantitative results on the COCO dataset. We
outperform the bottom-up OpenPose and even the top-down Mask R-CNN approach on all
metrics. These numbers are overall lower than their higher resolution counterparts. The
two conceptually very different baseline methods show similar performance while our method
is clearly ahead by over 18\% in AP.

\begin{figure*}
  \centering
  \includegraphics[width=1.0\linewidth, trim={0 700px 0 0}, clip]{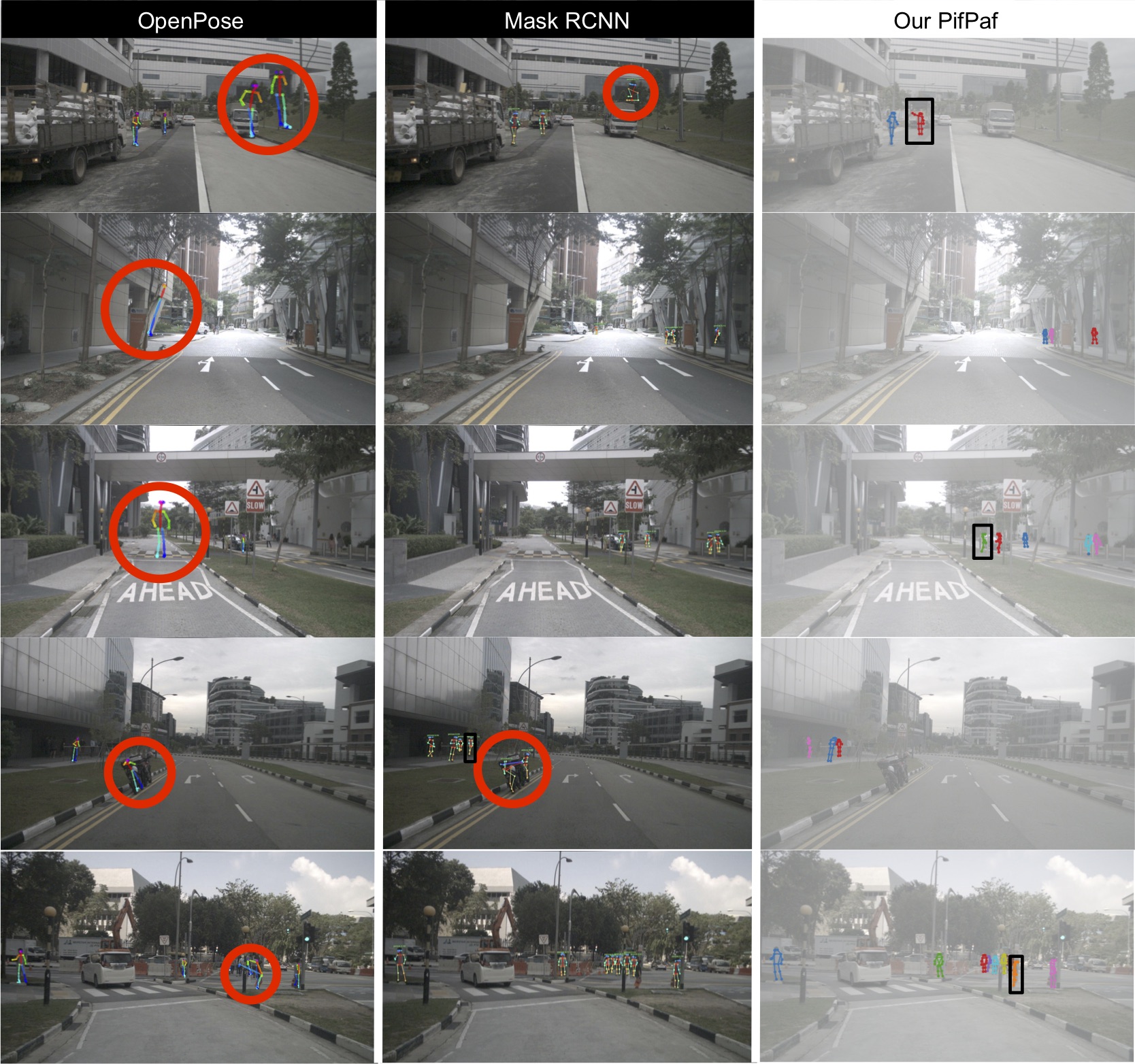}
  \caption{Illustration of our PifPaf method (right hand-side) against OpenPose~\cite{partsaffinityfields} (first column) and Mask R-CNN~\cite{tseng2018detectron} (second column) on the \textit{nuScenes} dataset. We highlight with bounding boxes all humans that other methods did not detect, and with circles all false positives. Note that our method correctly estimates the waving pose (first row, first bounding box) of a person whereas the others fail to do so.}
  \label{fig:all2}
\end{figure*}

Our quantitative results emulate the person distribution in urban street scenes
using a public, annotated dataset. Figure~\ref{fig:all2} shows
qualitative results of the kind of street scenes we want to address. Not only do we have
less false positives, we detect pedestrians who partially occlude each other. It is
interesting to see that a critical gesture such as ``waving'' towards a car is only
detected with our method. Both Mask-RCNN and OpenPose have not accurately estimated the
arm gesture in the first row of Figure~\ref{fig:all2}. Such level of difference can be
fundamental in developing safe self-driving cars.

\begin{figure*}
  \centering
  \includegraphics[width=1.0\linewidth, trim={0 780px 0 0}, clip]{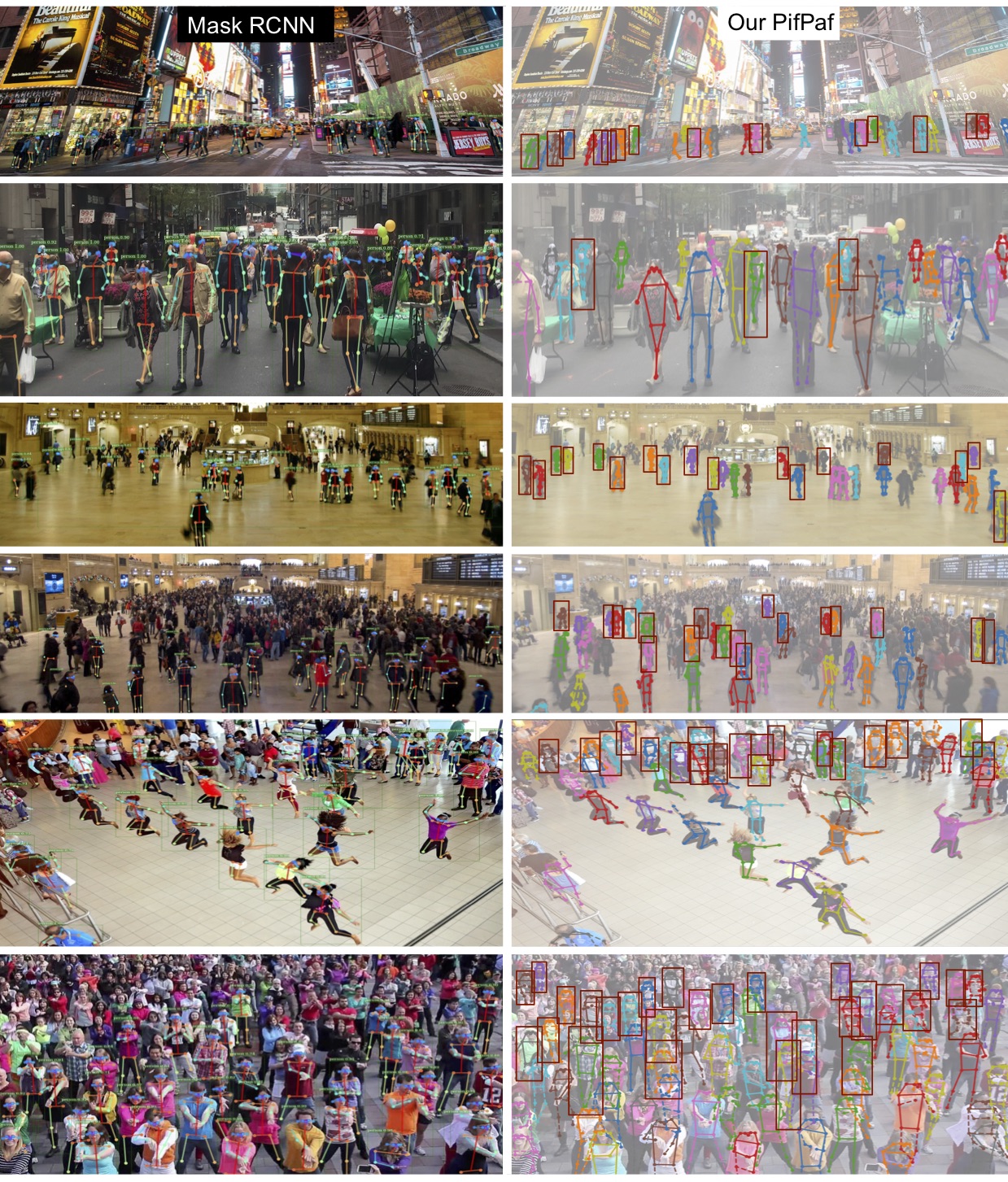}
  \caption{Illustration of our PifPaf method (right hand-side) against Mask R-CNN~\cite{tseng2018detectron} (left hand-side). We highlight with bounding boxes all humans where Mask R-CNN misses their poses with respect to our method. Our method estimates all poses that Mask R-CNN estimates as well as the ones highlighted with bounding boxes.}
  \label{fig:all}
\end{figure*}

We further show qualitative results on more crowded images in Figure~\ref{fig:all}. For
perspectives like the one in the second row, we observe that bounding boxes of close-by
pedestrians occlude further away pedestrians. This is a difficult scenario for top-down
methods. Bottom-up methods perform here better which we can also observe  for our PifPaf
method.

To quantify the performance on the Market-1501 dataset, we created a simplified accuracy
metric. The accuracy is 43\% for Mask R-CNN and 96\% for PifPaf.
The evaluation is based on the number of images with a correct
pose out of 202 random images from the train set.
A correct pose has up to three joints misplaced.

Other methods are optimized for higher resolution images. For a fair comparison,
we show a quantitative comparison on the high resolution COCO 2017 test-dev set
in Table~\ref{tab:high-res}. We perform on par with the best existing
bottom-up method.
\begin{table}
  \centering
  \begin{tabular}{|l|c c c|}

    \hline
                            & AP & AP$^{M}$ & AP$^{L}$ \\
    \hline\hline
    Mask R-CNN~\cite{he2017mask}            & 63.1 & 58.0 & 70.4 \\

    \hline
    OpenPose~\cite{partsaffinityfields}                 & 61.8 & 57.1 & 68.2 \\
    PersonLab~\cite{personlab} -- single-scale          & 66.5 & \textbf{62.4} & 72.3 \\

    \textbf{PifPaf} -- single-scale (ours)  & \textbf{66.7} & \textbf{62.4} & \textbf{72.9} \\
    \hline
  \end{tabular}
  \caption{
    Metrics in percent evaluated on the COCO 2017 test-dev set at optimal
    resolutions for top-down (top part) and bottom-up (bottom part) methods.
  }
  \label{tab:high-res}
\end{table}

\paragraph{Ablation Studies.}

We studied the effects of various design decisions that are summarized
in Table~\ref{tab:ablation}.
\begin{table}
  \centering
  \begin{tabular}{|l|c c c|}
    \hline
                                           & AP   & AP$^M$ & AP$^L$ \\
    \hline\hline
    vanilla $L_1$                          & 41.7 & 26.5   & 62.5 \\
    SmoothL1, $r=0.2\sqrt{A_i}\sigma_k$    & 42.0 & 26.9   & 62.6 \\
    SmoothL1, $r=0.5\sqrt{A_i}\sigma_k$    & 41.9 & 27.0   & 62.5 \\
    SmoothL1, $r=1.0\sqrt{A_i}\sigma_k$    & 41.6 & 26.5   & 62.3 \\

    Laplace                                & 45.1 & \textbf{31.4} & 64.0 \\
    Laplace (using $b$ in decoder)         & \textbf{45.5} & \textbf{31.4} & \textbf{64.9} \\
    \hline
  \end{tabular}
  \caption{Study of the dependence on the type of $L_1$ loss.
           Metrics are reported in percent.
           All models have a ResNet50 backbone and were trained for 20 epochs.}
  \label{tab:ablation}
\end{table}
We found that we can tune the performance
towards smaller or larger objects by modifying the overall scale of
$r^\textrm{smooth}$ and so we studied its impact. However, the real
improvement is obtained with the Laplace-based loss.
The added scale component $\sigma$ to the PIF field improved AP of our ResNet101
model from~64.5\%~to~65.7\%.

\paragraph{Runtime.}
Metrics for varying ResNet backbones
are in Table~\ref{tab:backbones}. For the same backbone, we outperform
PersonLab by 9.5\% in AP with a simultaneous 32\% speed up.
\begin{table}
  \centering
  \begin{tabular}{|l|c c c|}
    \hline
               & AP [\%] & $t$ [ms] & $t^{dec}$ [ms] \\
    \hline\hline
    ResNet50   & 62.6 & 222 & 178 \\
    ResNet101  & \textbf{65.7} (60.0) & \textbf{240} (355) & 175 \\
    ResNet152  & 67.4 & 263 & 173 \\
    \hline
  \end{tabular}
  \caption{Interplay between precision and single-image prediction time $t$ on a GTX1080Ti with different ResNet backbones for the COCO val set. Last column is the decoding time $t^{dec}$. PersonLab~\cite{personlab} timing numbers (which include decoding instance masks) are given in parenthesis where available at image width 801px.}
  \label{tab:backbones}
\end{table}

\section{Conclusions}

We have developed a new bottom-up method for multi-person 2D human pose estimation
that addresses failure modes that are particularly prevalent
in the transportation domain, \ie, in self-driving cars and social
robots. We demonstrated that our method outperforms previous state-of-the-art methods
in the low resolution regime and performs on par at high resolution.

The proposed PAF fields can be applied to other tasks as well. Within the image domain,
predicting structured image concepts~\cite{krishnavisualgenome} is an exciting next step.

\paragraph{Acknowledgements.}
We would like to thank EPFL SCITAS for their support with compute infrastructure.

\clearpage

{\small
\bibliographystyle{ieee}
\bibliography{references}
}

\end{document}

%% file: images/offset_grid.tex
\begin{tikzpicture}
    \filldraw[fill=green!50!white, draw=green!20!white] (-1.0,-1.0) rectangle (-0.5,-0.5);
    \draw[step=0.5, gray, very thin] (-1.5,-1.5) grid (1.5,1.5);

    \filldraw[fill=blue!50!white, draw=blue!50!white] (1.2,0.4) circle [radius=0.1];
    \filldraw[fill=blue!50!white, draw=blue!50!white] (-1.2,0.7) circle [radius=0.1];
    \filldraw[fill=blue!50!white, draw=blue!50!white] (-0.625,-0.825) circle [radius=0.1];
    \filldraw[fill=blue!50!white, draw=blue!50!white] (-0.825,-0.625) circle [radius=0.1];

    \draw[->, thick] (-0.75,-0.75) -- (1.2,0.4);
  \end{tikzpicture}

%% file: images/paf_grid.tex
\begin{tikzpicture}
    \filldraw[fill=green!40!white, draw=green!40!white] (0.0,-1.0) rectangle (0.5,-0.5);
    \filldraw[fill=green!50!white, draw=green!50!white] (-1.0,-0.5) rectangle (-0.5,0.0);

    \draw[step=0.5, gray, very thin] (-1.5,-1.5) grid (1.5,1.5);

    \filldraw[fill=blue!50!white, draw=blue!50!white] (1.2,0.4) circle [radius=0.1];
    \filldraw[fill=blue!50!white, draw=blue!50!white] (-1.2,0.7) circle [radius=0.1];
    \filldraw[fill=blue!50!white, draw=blue!50!white] (-0.625,-0.825) circle [radius=0.1];
    \filldraw[fill=blue!50!white, draw=blue!50!white] (-0.825,-0.625) circle [radius=0.1];


    \draw[->, thick] (0.25,-0.75) -- (-0.625,-0.825);
    \draw[->, thick] (0.25,-0.75) -- (1.2,0.4);


    \draw[->, thick] (-0.75,-0.25) -- (-0.825,-0.625);
    \draw[->, thick] (-0.75,-0.25) -- (-1.2,0.7);

  \end{tikzpicture}